\documentclass[runningheads]{llncs}
\usepackage{comment}
\usepackage{array}
\usepackage{booktabs}
\usepackage{graphicx}
\usepackage{subfigure}
\usepackage{multirow}
\usepackage{comment}
\usepackage{adjustbox}
\usepackage{color}
\usepackage{graphicx}
\usepackage{comment}
\usepackage{amsmath,amssymb}
\usepackage[switch]{lineno}

\DeclareMathOperator*{\argmax}{argmax}

\def\P{{\mathbf P}}
\newcommand{\RNum}[1]{\uppercase\expandafter{\romannumeral #1\relax}}
\begin{document}
\title{3D Hand Pose Estimation via Regularized Graph Representation Learning \thanks{This work was supported by National Natural Science Foundation of China under contract No. 61972009.} \fnmsep \thanks{Corresponding author: Wei Hu (forhuwei@pku.edu.cn)}}
\def\YOFOSubNumber{228}  

\titlerunning{Hand Pose Estimation via Graph}
%
\author{Yiming He\inst{1} \and
Wei Hu\inst{1}}
\authorrunning{Yiming and Wei}
%
\institute{Wangxuan Institute of Computer Technology, Peking University }
\maketitle              

\begin{abstract}
This paper addresses the problem of 3D hand pose estimation from a monocular RGB image. 
While previous methods have shown great success, the structure of hands has not been fully exploited, which is critical in pose estimation.   
To this end, we propose a regularized graph representation learning under a conditional adversarial learning framework for 3D hand pose estimation, aiming to capture structural inter-dependencies of hand joints. 
In particular, we estimate an initial hand pose from a parametric hand model as a prior of hand structure, which regularizes the inference of the structural deformation in the prior pose for accurate graph representation learning via residual graph convolution. 
To optimize the hand structure further, we propose two bone-constrained loss functions, which characterize the morphable structure of hand poses explicitly. 
Also, we introduce an adversarial learning framework conditioned on the input image with a multi-source discriminator, which imposes the structural constraints onto the distribution of generated 3D hand poses for anthropomorphically valid hand poses. 
Extensive experiments demonstrate that our model sets the new state-of-the-art in 3D hand pose estimation from a monocular image on five standard benchmarks.

\end{abstract}

\vspace{-0.3in}
\section{Introduction}
\label{sec:intro}
3D human hand pose estimation is a long-standing problem in computer vision, which is critical for various applications such as virtual reality and augmented reality \cite{Hrst2013,Piumsomboon13}.
Previous works attempt to estimate hand pose from depth images \cite{Ge2016Robust,handpointnet} or in multi-view setups \cite{Panteleris2017Back,Zhang20163D}. 
However, due to the diversity and complexity of hand shape, gesture, occlusion, {\it etc.}, it still remains a challenging problem despite years of studies \cite{Hui2017Hough}.

As RGB cameras are more widely accessible than depth sensors, recent works focus mostly on 3D hand pose estimation from a monocular RGB image and have shown their efficiency \cite{Baseline,Wild,Pushing,Cai_2018_ECCV,Doosti_2020_CVPR}.
While some early works \cite{Cai_2018_ECCV,Wild} did not explicitly exploit the structure of hands, some recent methods \cite{Baseline,Doosti_2020_CVPR} have shown the crucial role of hand structure in pose estimation, but may resort to an additional synthetic dataset. 
Also, unlike bodies and faces that have obvious local characteristics ({\it e.g.}, eyes on a face), hands exhibit almost uniform appearance. 
Consequently, estimated hand poses from existing methods are sometimes distorted and unnatural.

\begin{figure}[t]
     \centering
     \includegraphics[width=0.7\linewidth]{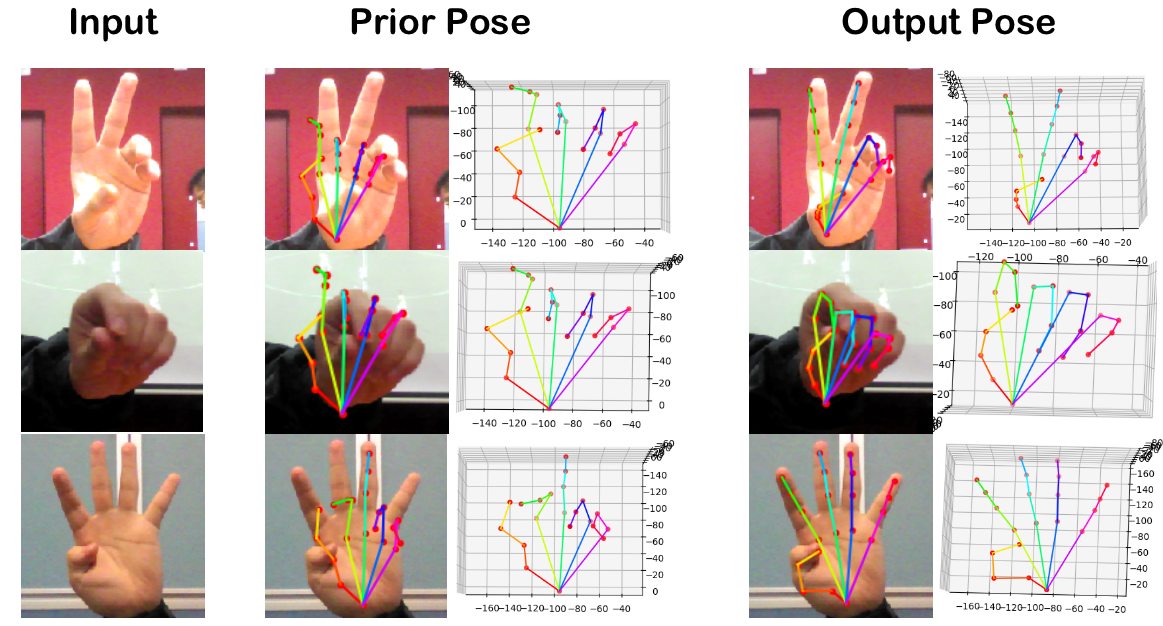}
     \vspace{-0.2in}
     \caption{\small{\textbf{The proposed method estimates 3D hand pose from a monocular image based on regularized graph representation learning.} 
     A parametric hand model generates a \textit{prior pose}, which regularizes the learning of deformations in graph topology under a conditional adversarial learning framework.
     }}
     \vspace{-0.25in}
     \label{teaser}
\end{figure}



To fully exploit the structure of hands, we propose to represent the irregular topology of 3D hand poses naturally on graphs, and learn the graph representation regularized by a prior pose from the monocular image input under a conditional generative adversarial learning framework, aiming to capture the structural dependencies among hand joints.
Moreover, while most existing works \cite{Wild,Baseline,Cai_2018_ECCV} deploy 3D Euclidean distance between joints as the loss function for 3D annotation, we propose two {\it bone loss functions} that constrain the length and orientation of each bone connected by adjacent joints so as to preserve hand structure explicitly.
Besides, unlike some recent works \cite{Baseline,Cai_2018_ECCV,dkulon2019rec}, we estimate 3D hand poses {\it without} resorting to ground truth meshes or depth maps, which is more suitable for datasets in the wild.

Specifically, given an input monocular image, our framework consists of a hand pose generator and a conditional discriminator. 
The generator is composed of a MANO hand model module \cite{MANO} that provides an initial pose estimation as prior pose and a deformation learning module regularized by the prior pose.  
In particular, taking the prior pose and image features as input, the deformation learning module learns the deformation in the prior pose to further refine the hand structure, by our designed residual graph convolution that leverages on the recently proposed ResGCN \cite{graphres}.
Further, we design a conditional multi-source discriminator that employs hand poses, hand bones computed from poses as well as the input image to distinguish the predicted 3D hand pose from the ground-truth, leading to anthropomorphically valid hand pose. 
Experimental results demonstrate that our model achieves significant improvements over state-of-the-art approaches on five standard benchmarks.

To summarize, our main contributions include 
\begin{itemize}
    \item We propose regularized graph representation learning for 3D hand pose estimation from a monocular image, which fully exploits structural information. 
    
    \item We learn the graph representation of hand poses by inferring structural deformation, which is regularized by an initial hand pose estimation from a parametric hand model.
    
    \item We introduce two bone-constrained loss functions, which optimize the estimation of hand structures by explicitly enforcing constrains on the topology of bones.
    
    \item We present a conditional adversarial learning framework to impose structural constraints onto the distribution of generated 3D hand poses, which is able to address the challenge of uniform appearance in hands. 
    
    
    
\end{itemize}


\vspace{-0.3in}
\section{Related Work}
\label{sec:related}
According to the input modalities, previous works on 3D hand pose estimation can be classified into two categories: 
1) 3D hand pose estimation from depth images; 
2) 3D hand pose estimation from a monocular RGB image.

\vspace{-0.1in}
\subsection{Estimation from Depth Images}
\vspace{-0.1in}
Depth images contain rich 3D information for hand pose estimation \cite{Tang2013Real}, which has shown promising accuracy \cite{Yuan2017Depth}. 
There is a rich literature on 3D hand pose estimation with depth images as input \cite{handpointnet,Ge2016Robust,Fitzgibbon2015Accurate,Choi2016DeepHand,De2011Model,Khamis2015Learning,Malik2018DeepHPS}. 
Among them, some earlier works such as  \cite{De2011Model,Khamis2015Learning} are based on a deformable hand model with an iterative optimization training approach. Due to the effectiveness of deep learning, some recent works like \cite{Malik2018DeepHPS}
leverage CNN to learn the shape and pose parameters for a proposed model (LBS hand model).


\vspace{-0.1in}
\subsection{Estimation from a Monocular Image}
\vspace{-0.1in}
Compared with the aforementioned two categories, a monocular RGB image is more accessible. 
Early works \cite{Athitsos2003Estimating} propose complex model-fitting approaches, which are based on dynamics and multiple hypotheses and depend on restricted requirements. 
These model-fitting approaches have proposed many hand models, based on assembled geometric primitives \cite{model1} or sphere meshes \cite{model2}, {\it etc.} 
Our work deploys the MANO hand model \cite{MANO} as our prior, which models both hand shape and pose as well as generates meshes. 
Nevertheless, these sophisticated approaches suffer from low estimation accuracy.

With the advance of deep learning, many recent works estimate 3D hand pose from a monocular RGB image using neural networks \cite{Baseline,Wild,Pushing,Cai_2018_ECCV,dkulon2019rec}. 
Among them, some recent works \cite{dkulon2019rec,Baseline} directly reconstruct the 3D hand mesh and then generate the 3D hand pose through a pose regressor.
Kulon {\it et al.} \cite{dkulon2019rec} reconstruct the hand pose based on an auto-encoder, which employs an encoder to extract the latent code and feeds the latent code into the decoder to reconstruct hand mesh.
Ge {\it et al.} \cite{Baseline} propose to estimate vertices of 3D meshes from GCNs \cite{GCN} in order to learn nonlinear variations in hand shape. 
The latent feature of the input RGB image is extracted via several networks and then fed into a GCN to directly infer the 3D coordinates of mesh vertices. 
However, since the accuracy of the output hand mesh is critical for both methods, they need an extra dataset which provides ground truth hand meshes as supervision.
Also, the upsampling layer used in \cite{Baseline} to reconstruct the hand mesh will cause a non-uniform distribution of vertices in mesh, which influences the accuracy of hand pose.

In contrast, we take a prior pose estimated from a parametric hand model as regularization for graph representation learning over hand poses rather than directly reconstructing hand poses from latent features. 
Besides, our method does not require any additional supervision such as mesh supervision \cite{Baseline,dkulon2019rec} or depth image supervision \cite{Baseline,Cai_2018_ECCV}. 
Hence, our method is more suitable for datasets in the wild.
Further, we introduce conditional adversarial training for 3D hand pose estimation, which enables learning a real distribution of 3D hand poses.

\vspace{-0.2in}
\section{Methodology}
\label{sec:overview}
\subsection{Overview of the Proposed Approach}
\begin{figure*}[t]
	\centering
	\includegraphics[width=0.65\linewidth]{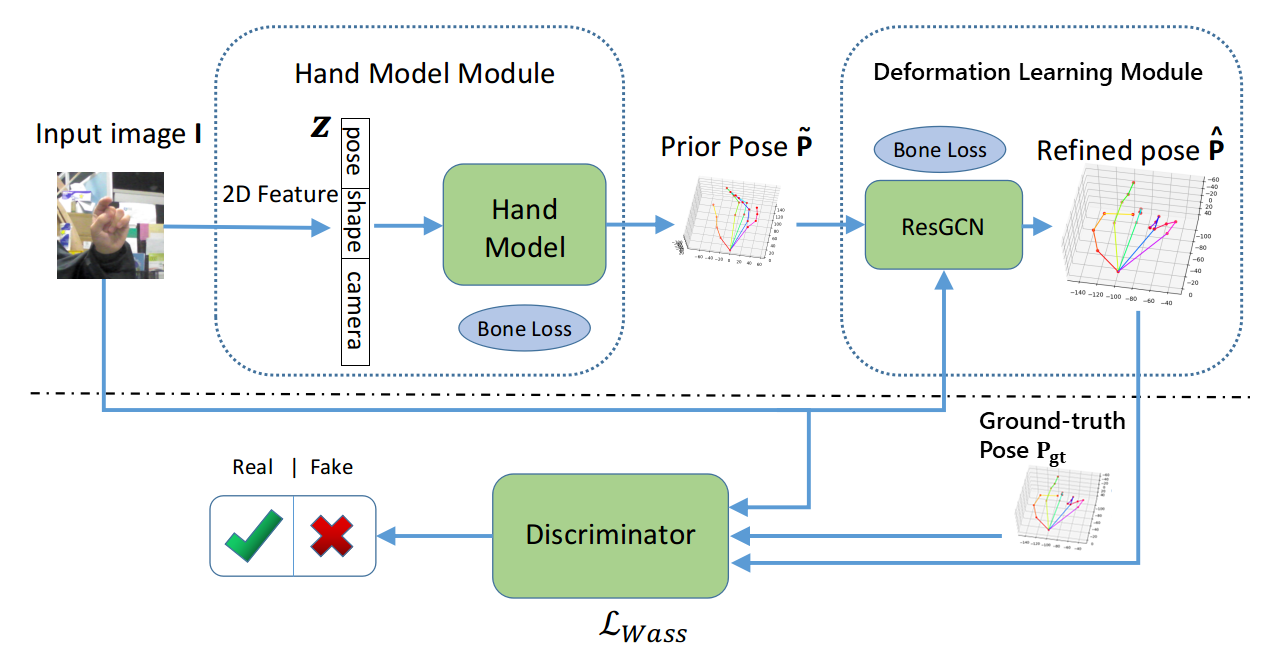}
	\vspace{-0.1in}
	\caption{{Architecture of the proposed regularized graph representation learning under a conditional adversarial learning framework for 3D hand pose estimation.}}
	\label{fig:overview}
\end{figure*}
We aim to infer 3D hand pose via regularized graph representation learning under an adversarial learning framework. 
The entire framework consists of a hand pose generator $\mathbb{G}$ and a conditional discriminator $\mathbb{D}$, as illustrated in Fig.~\ref{fig:overview}.

	

The multi-source discriminator $\mathbb{D}$ imposes structural constraints onto the distribution of generated 3D hand poses conditioned on the input image, which distinguishes the ground-truth 3D poses from the predicted ones. 


\subsection{The Proposed Hand Pose Generator $\mathbb G$}

Given the observed input image $\mathbf{I}$ and ground truth hand pose $\P_{\text{gt}}$, we formulate the training of hand pose estimation from a monocular image as a Maximum a Posteriori (MAP) estimation problem: 
\begin{equation}
    \hat{\P}_{\text{MAP}}(\mathbf{I},\P_{\text{gt}}) = \argmax_{\P} f(\mathbf{I},\P_{\text{gt}}|\P) g(\P),
    \label{eq:MAP}
\end{equation}
where $\P$ denotes the hand pose to estimate. 
In \eqref{eq:MAP}, $g(\P)$ represents the prior probability distribution of the hand pose, which provides the prior knowledge of $\P$. 
$f(\mathbf{I},\P_{\text{gt}}|\P)$ denotes the likelihood function, which is the probability of obtaining the observed image $\mathbf{I}$ and ground truth hand pose $\P_{\text{gt}}$ given the estimated hand pose $\P$.

We define the likelihood function as an exponential function of the distance between the estimated pose and the ground truth pose/input image:
\begin{equation}
    f(\mathbf{I},\P_{\text{gt}}|\P) = \exp\{-d_1(\P_{\text{gt}},\P)-d_2(\mathbf{I},\P)\},
    \label{eq:likelihood}
\end{equation}
where $d_1(\cdot)$ is the distance metric between the estimated hand pose and the ground truth, and $d_2(\cdot)$ is the distance metric between the estimated hand pose and the input image. 
Regarding $g(\P)$, it is a constant $C$ after we acquire a prior pose from a parametric hand model. 
Hence, when we substitute \eqref{eq:likelihood} and $g(\P)=C$ into \eqref{eq:MAP}, take the logarithm and multiply by $-1$, we have 
\begin{equation}
    \min_{\P} d_1(\P_{\text{gt}},\P)+d_2(\mathbf{I},\P).
    \label{eq:objective}
\end{equation}
$d_1(\cdot)$ and $d_2(\cdot)$ will be discussed in Section~\ref{sec:boneloss} in detail. 

Specifically, we employ a parametric hand model to provide the prior of $\P$, and designate a Deformation Learning Module to learn the pose under the supervision of the ground-truth pose and input image.
We discuss the two modules of the generator in detail as follows.

\subsubsection{The Hand Model Module}
Given an input monocular image, this module aims to generate an initial estimation of 3D hand pose $\Tilde{\mathbf{P}}$ as a prior. 
A hand model is able to represent both hand shape and pose with a few parameters, which is thus a suitable prior for hand pose estimation. 



We first predict parameters of the hand model. 
Specifically, we crop and resize the input image to a salient region of the hand, which is fed into the ResNet-50 network \cite{ResNet} to extract features for the construction of the latent code $\mathbf{z}$, {\it i.e.}, parameters of the hand model.  
Then, we employ a modified MANO hand model \cite{MANO}, which is based on the SMPL model \cite{SMPL} for human bodies. 

\subsubsection{The Deformation Learning Module}

\label{GB}
\begin{figure}[t]
    \centering
    \includegraphics[width=0.7\linewidth]{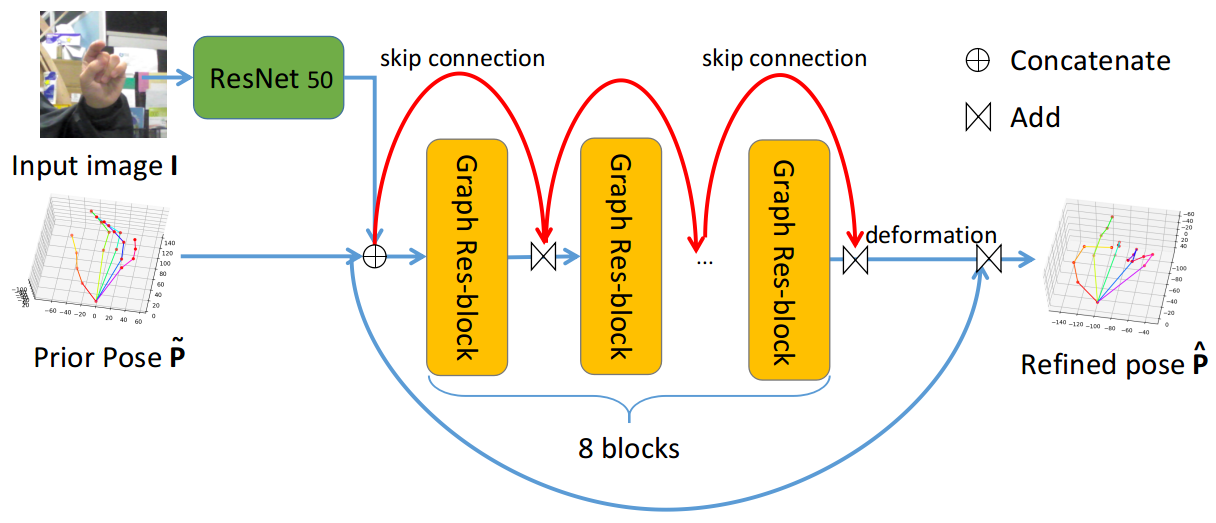}
    \vspace{-0.1in}
    \caption{\small{{Architecture of the deformation learning module in our generator.}}}
    \label{posegcn}
\end{figure}

This module aims at accurate graph representation learning for hand pose estimation, which is conditional on the prior and under the supervision of the input image and ground truth pose as in \eqref{eq:MAP}. 
In particular, conditioned on the prior $\tilde{\mathbf{P}}$, we learn the structural {\it deformation} in $\tilde{\mathbf{P}}$ instead of the holistic hand pose.   

We first construct an unweighted graph over $\Tilde{\mathbf{P}}$, where the irregularly sampled key points ({\it i.e.}, joints) on the hand are projected onto nodes. 
The graph signal on each node is the concatenation of the global feature vector of the input image and the 3-dimensional coordinate vector of each joint in the input prior pose. 
Nodes are connected if they represent adjacent key points of the hand, where the adjacency relations follow the human hand structure as presented in Fig.~\ref{hands}, leading to an adjacency matrix $\mathbf{A} \in \mathbb{R}^{N \times N}$.  

Based on the graph representation $\mathbf{A}$, the finally refined pose is 
\begin{equation}
\hat{\mathbf{P}} = \Tilde{\mathbf{P}} + \mbox{GCN}(\Tilde{\mathbf{P}}\oplus \mathbf{F},\mathbf{A}),
 \label{eq5}
 \end{equation}
where $\mathbf{F} \in \mathbb{R}^{N \times F}$ denotes the $F$-dimensional global feature vector of the image repeated $N$ times, and $\oplus$ denotes the feature-wise concatenation operation. 
$\mbox{GCN}(\Tilde{\mathbf{P}} \oplus \mathbf{F},\mathbf{A})$ represents the learned deformation between the prior $\Tilde{\mathbf{P}}$ and the ground truth. 
The sum of the prior pose $\Tilde{\mathbf{P}}$ and its deformation thus leads to the refined hand pose.  


Let $\mathbf X^l$ denote the input of the $l$-th Graph Res-block, then the output of the $l$-th Graph Res-block takes the form
\begin{equation}
    \mathbf{X}^{l+1} =  N\left(g(N(g(\mathbf{X}^l,\mathbf{A})),\mathbf{A})\right) + \text{skip}(\mathbf{X}^l), 
    \label{eq:resblock}
\end{equation}
where $g(\cdot)$ represents a single GCN layer as in \cite{GCN}, $N(\cdot)$ represents a single normalization layer, and $\text{skip}(\cdot)$ denotes the skip connection which is a GCN layer to match the dimension of the two terms in \eqref{eq:resblock}.  
We then stack several layers of Graph Res-blocks to learn the deformation of the prior pose, as demonstrated in Fig.~\ref{posegcn}. 

\subsection{The Proposed Conditional Discriminator $\mathbb D$} \label{sec:discriminator}
\begin{figure}[t]
	\centering
	\includegraphics[width=0.7\linewidth]{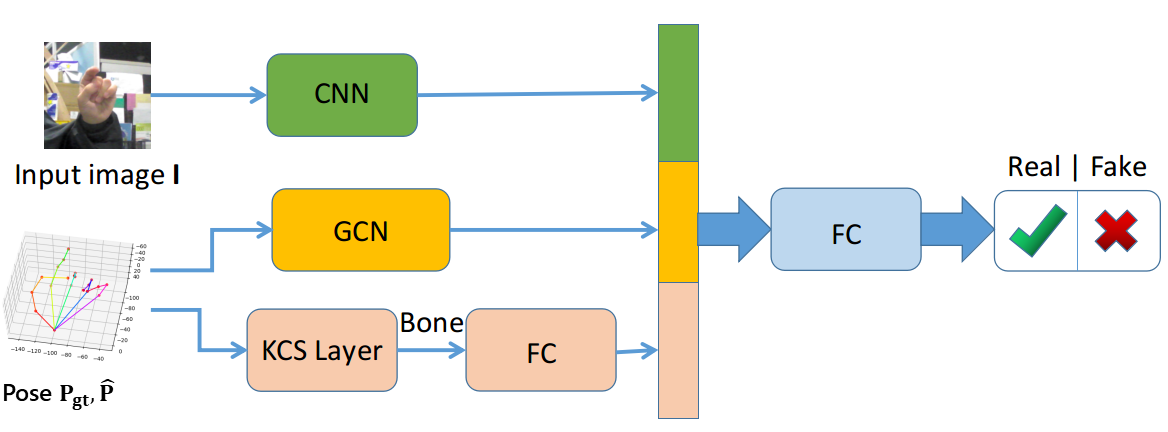}
	\vspace{-0.1in}
	\caption{\small{{Architecture of the conditional discriminator.} 
	}}
	\label{discriminator}
\end{figure}


A simple architecture of a discriminator is a fully-connected (FC) network with the hand pose as input, which however has two shortcomings: 
1) the relation between the RGB image and inferred hand pose is neglected; 
2) structural properties of the hand pose are not taken into account explicitly.
Instead, inspired by the multi-source architecture in \cite{Multi-source}, we design a conditional multi-source discriminator with three inputs to address the aforementioned issues. 
As illustrated in Fig.~\ref{discriminator}, the inputs include: 
1) features of the input monocular image;
2) features of the refined hand pose $\hat{\mathbf{P}}$ or the ground truth pose $\P_{\text{gt}}$; 
3) features of bones via the KCS layer as in \cite{KCS}, which computes the bone matrix from $\hat{\mathbf{P}}$ or $\P_{\text{gt}}$ via a simple matrix multiplication.
The bone features contain prominent structural information such as the length and direction of bones, thus characterizing the hand structure accurately. 

The loss function of the conditional discriminator follows the definition of the Wasserstein loss \cite{WGAN} conditioned on the input image $\mathbf{I}$:
\begin{equation}
    \mathcal{L}_{\text{Wass}} = - \mathbb{E}_{\mathbf{P}_{\text{gt}}\sim p_{data}(\mathbf{P}_{\text{gt}})}\mathbb D(\mathbf{P}_{\text{gt}}|\mathbf{I}) + \mathbb{E}_{\hat{\mathbf{P}}\sim p(\hat{\mathbf{P}})}\mathbb D(\hat{\mathbf{P}}|\mathbf{I}),
    \label{eq:L_wass}
\end{equation}
where $\mathbb{D}$ takes the generated (fake) pose $\hat{\mathbf{P}}$ and ground-truth pose $\mathbf{P}_{\text{gt}}$ as input, $\mathbf{P}_{\text{gt}}$ is a sample following the ground-truth pose distribution $p_{data}(\mathbf{P}_{\text{gt}})$ and $\hat{\mathbf{P}}$ is a sample from the refined pose distribution $p(\hat{\mathbf{P}})$.


\subsection{The Proposed Bone-Constrained Loss Functions}
\label{sec:boneloss}
\begin{figure}[t]
    \centering
	\includegraphics[width=0.4\linewidth]{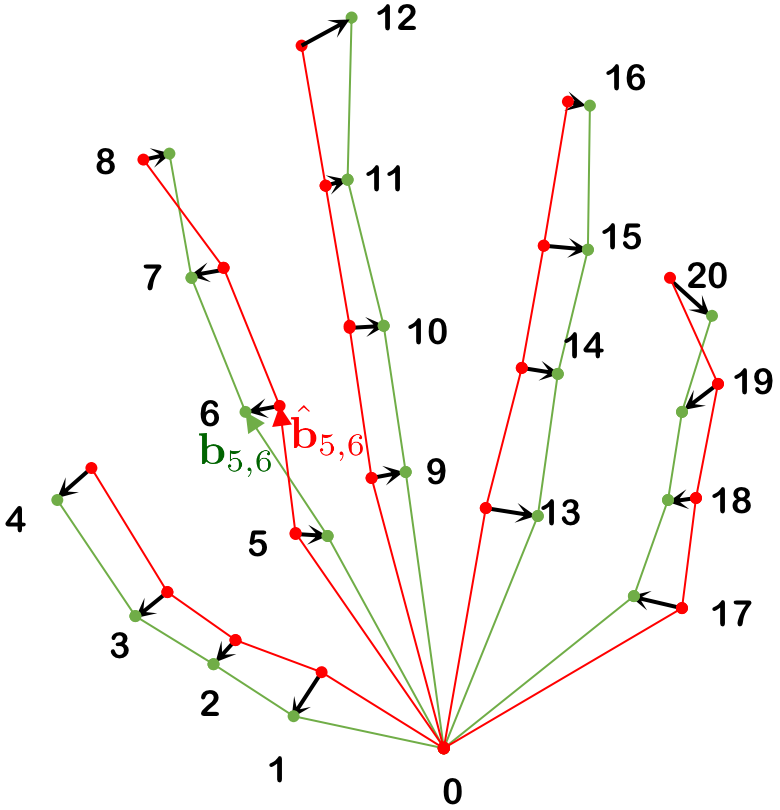}
	\vspace{-0.1in}
	\caption{\small{\textbf{Illustration of the residual between the ground truth hand pose (marked in green) and the predicted one (marked in red).}}
	Each hand pose has 21 key joints. 
	We denote a bone vector connecting two key joints $i$ and $j$ by $\mathbf{b}_{i,j}$, such as $\mathbf{b}_{5,6}$ in the figure.}
	\label{hands}
\end{figure}

As presented in \eqref{eq:objective}, we have two types of loss functions for the MAP estimation of hand pose. We employ the commonly adopted Euclidean distance in the coordinates of joints of 3D hand pose $\mathcal{L}_{\text{pose}}$ \cite{Baseline} as well as two proposed bone-constrained metrics as $d_1(\cdot)$ to measure the distortion of the estimated 3D hand pose compared to the ground truth, and apply the commonly used Euclidean distance in the coordinates of joints of projected 2D hand pose $\mathcal{L}_{\text{proj}}$ \cite{Baseline} as $d_2(\cdot)$ to measure the distance between the estimation and the 2D image,
\begin{equation}
    \mathcal{L}_{\text{pose}}=\sum_{i}||\mathbf{j}_i-\hat{\mathbf{j}}_i||_2, \mathcal{L}_{\text{proj}}=\sum_{i}||\mathbf{j'}_i-\hat{\mathbf{j'}}_i||_2,
\end{equation}
where $\mathbf{j}_i\in \mathbb{R}^{3 \times 1},\mathbf{j'}_i\in \mathbb{R}^{2 \times 1}$ are 3D and 2D coordinates of joint $i$ respectively. 

Since $\mathcal{L}_{\text{pose}}$ and $\mathcal{L}_{\text{proj}}$ cannot capture the structural properties of hand pose explicitly, we propose two novel bone-constrained loss functions to characterize the length and direction of each bone. 

As illustrated in Fig.~\ref{hands}, we first define a bone vector $\mathbf{b}_{i,j} \in \mathbb{R}^{3 \times 1}$ between hand joint $i$ and $j$ as \begin{equation}
    \mathbf{b}_{i,j}=\mathbf{j}_{i}-\mathbf{j}_{j},
\end{equation}   

The first bone-constrained loss $\mathcal{L}_{\text{len}}$ quantifies the distance in \textit{bone length} between the ground truth hand and its estimate, which we define as
\begin{equation}
    \mathcal{L}_{\text{len}}=\sum_{i,j}\left| ||\mathbf{b}_{i,j}||_2 - ||\hat{\mathbf{b}}_{i,j}||_2 \right|,
\end{equation}
where $\mathbf{b}_{i,j}$ and $\hat{\mathbf{b}}_{i,j}$ are the bone vectors of the ground truth and the predicted bone respectively. 

The second bone-constrained loss $\mathcal{L}_{\text{dir}}$ measures the deviation in the \textit{direction of bones}: 
\begin{equation}
  \mathcal{L}_{\text{dir}}=\sum_{i,j}\left|\left|\mathbf{b}_{i,j}/||\mathbf{b}_{i,j}||_2-\hat{\mathbf{b}}_{i,j}/||\hat{\mathbf{b}}_{i,j} ||_2\right|\right|_2. 
\end{equation}

Besides, as we adopt the framework of adversarial learning, we also introduce the Wasserstein loss $\mathcal{L}_{\text{Wass}}$ in \eqref{eq:L_wass} into the loss function for adversarial training. Hence, the overall loss function $\mathcal{L}$ is 
\begin{equation}
    \mathcal{L} = \mathcal{L}_{\text{pose}} + \lambda_{\text{proj}}\mathcal{L}_{\text{proj}} + \lambda_{\text{len}}\mathcal{L}_{\text{len}} + \lambda_{\text{dir}}\mathcal{L}_{\text{dir}} + 
    \lambda_{\text{Wass}}\mathcal{L}_{\text{Wass}},
    \label{eq:loss}
\end{equation}
where $\lambda_{\text{proj}}$, $\lambda_{\text{len}}$, $\lambda_{\text{dir}}$ and $\lambda_{\text{Wass}}$ are hyperparameters for the trade-off among these losses. 
In accordance with \eqref{eq:objective}, $d_1 = \mathcal{L}_{\text{pose}}  + \lambda_{\text{len}}\mathcal{L}_{\text{len}} + \lambda_{\text{dir}}\mathcal{L}_{\text{dir}}$, and $d_2 = \lambda_{\text{proj}}\mathcal{L}_{\text{proj}}$.

\vspace{-0.1in}
\section{Experimental Results}
\label{sec:results}

\vspace{-0.1in}
\subsection{Implementation Details}
In our experiments, we first pretrain the hand model module and then train the entire network end-to-end. In particular, the training process can be divided into three stages.

\textbf{Stage \RNum{1}.} We pretrain the hand model module, which is randomly initialized and trained for 100 epochs using the Adam optimizer with learning rate 0.001. Then, we freeze the parameters of this stage to evaluate the effectiveness of the deformation learning module.

\textbf{Stage \RNum{2}.} We train the generator $\mathbb{G}$ end-to-end without the discriminator $\mathbb{D}$. 
In $\mathbb{G}$, the hand model module is initialized with the trained model in the first stage and the deformation learning module is randomly initialized. 
$\mathbb{G}$ is then trained with 100 epochs using the Adam optimizer with learning rate 0.0001. 

\textbf{Stage \RNum{3}.} We adopt the framework of SNGAN \cite{SNGAN} for the conditional adversarial training, and train our model end-to-end.
$\mathbb{G}$ and $\mathbb{D}$ are trained with 100 epochs using the Adam optimizer with learning rate 0.0001. 

Regarding the hyper-parameters in \eqref{eq:loss}, we set $\lambda_{\text{len}}=0.01,\lambda_{\text{dir}}=0.1,\lambda_{\text{proj}}=0.1, \lambda_{\text{Wass}}=0.01$.



\vspace{-0.1in}
\subsection{Experimental Results}
\vspace{-0.1in}
\begin{table}[t]
\centering
\setlength{\tabcolsep}{2.5mm}{
\begin{tabular}{l|l|l|l|l|l}
\toprule[1pt]
             & STB  & RHD   & MPII+ZNSL(px) & Dexter+Object & EgoDexter \\ \midrule[0.5pt]
\cite{Baseline}     & 6.37 & 15.33 & -             & -             & -         \\ \hline
\cite{Wild}      & 9.76 & -     & 18.95         & 25.53         & 45.33     \\ \hline
\cite{Spurr}     & 8.56 & 19.73 & -             & 40.20         & 56.92     \\ \hline
\cite{Zimmermann_2017_ICCV} & -    & -     & 59.40         & 34.75         & 52.77     \\ \hline
Ours         & \textbf{3.97} & \textbf{12.40} & \textbf{9.87}          & \textbf{16.12}         & \textbf{34.98}     \\
\bottomrule[1pt]
\end{tabular}}
\caption{\small{{Comparison with state-of-the-art methods on the five datasets.} Note that MPII+ZNSL only provides 2D annotation, thus we employ the 2D distance (px) metric on this dataset.}}
\label{tab:result}
\vspace{-0.2in}
\end{table}

We compare our method with competitive 3D hand pose estimation approaches on the five datasets. 
We list the results in 3D Euclidean distance for comparison with the state-of-the-arts in Tab.~\ref{tab:result}. 
Compared to these works which directly reconstruct the 3D hand pose \cite{Baseline,Wild,Cai_2018_ECCV}, our method performs much better mainly due to the proposed regularized graph representation learning and conditional adversarial learning. 
We show the qualitative results and PCK results in the supplementary material. 





\vspace{-0.2in}
\subsection{Ablation Studies}
\begin{table}[t]
    \centering
    \begin{minipage}{\linewidth}
    \centering
    \setlength{\tabcolsep}{1.5mm}{
    \begin{tabular}{c|ccc|ccc}
    \toprule[1pt]
    Stage & hand model & deformation & discriminator & STB     & RHD & EGODEXTER     \\ \midrule[0.5pt]
    \RNum{1}     & \checkmark          &            &               & 24.15 & 83.37 & 52.32 \\
    \RNum{2}     & \checkmark          & \checkmark          &               & 5.12  & 15.84 & 43.26 \\
    \RNum{3}     & \checkmark          & \checkmark          & \checkmark             & \textbf{3.97}  & \textbf{12.40} & \textbf{34.98}\\ \bottomrule[1pt]
    \end{tabular}}
    \caption{\small{{The performance of different stages in our model on three datasets (measured in 3D Euclidean distance (mm)).}}}
    \label{table1}
    \vspace{-0.1in}
    \end{minipage} \\[3pt]
    \begin{minipage}{\linewidth}
    \centering
    \begin{tabular}{c|ccc|ccc}
    \toprule[1pt]
    Model & GCN Deformation & FC Deformation & Discriminator & STB     & RHD  & EGODEXTER   \\ \midrule[0.5pt]
    1     &                & \checkmark             &                            & 15.11 & 37.59 & 52.34 \\
    2     & \checkmark              &               &                            & 5.12  & 15.84 & 40.12 \\
    3     &                & \checkmark             & \checkmark                          & 10.23 & 25.15 & 44.23 \\ 
    4     & \checkmark              &               & \checkmark                          & \textbf{3.97}  & \textbf{12.40} & \textbf{34.98} \\\bottomrule[1pt]
    \end{tabular}
    \caption{\small{Ablation studies on the Deformation Learning Module, with comparison between the Deformation Learning Module and the simple FC Refinement Module in 3D Euclidean distance (mm).}}
    \label{tab:refinement}
    \vspace{-0.4in}
    \end{minipage}
\end{table}
We perform ablation studies on the performance of different stages, the deformation learning module, the discriminator and loss functions. Due to the page limit, we present all the results in 3D Euclidean distance (mm). Please refer to the supplementary material for the results measured in 3D PCK. 

\textbf{On different stages.}
We present the results of three training stages in average 3D Euclidean distance, as listed in Tab.~\ref{table1}. 
The performance of {\bf Stage \RNum{2}} significantly outperforms {\bf Stage \RNum{1}}, which demonstrates that the proposed deformation learning module plays the most critical role in our model. 
The adversarial training scheme (\textbf{Stage \RNum{3}}) further improves the result, by learning a real distribution of the 3D hand pose. 

\textbf{On the deformation learning module.}
We compare the deformation learning module with a simple fully-connected deformation learning module (FC Deformation Module) to refine the prior pose. 
We train the deformation learning modules in different experimental settings: 1) without our discriminator, {\it i.e.}, without adversarial learning; and 2) with our discriminator. As presented in Tab.~\ref{tab:refinement}, the GCN deformation learning module leads to significant gain over the simple FC deformation module on both datasets in different settings, thus validating the superiority of the proposed deformation learning module.

\textbf{On the conditional discriminator.}
We compare with a single-source discriminator which only takes the 3D hand pose as the input. 
As presented in Tab.~\ref{tab:discriminator}, the multi-source discriminator outperforms the single-source one on both datasets, which gives credits to exploring the structure of hand bones and the relation between the image and pose.

\begin{table}[t]
\centering
\begin{adjustbox}{width=\linewidth}
\begin{tabular}{c|ccc|ccc}
\toprule[1pt]
Model & Deformation Learning & Multi-source & Single-source  & STB    & RHD  & EGODEXTER   \\ \midrule[0.5pt]
1     & \checkmark          & \checkmark                          &                             & \textbf{3.97} & \textbf{12.40} & \textbf{34.98} \\ 
2     & \checkmark          &                            & \checkmark                           & 4.54 & 15.10 & 37.46 \\ \bottomrule[1pt]
\end{tabular}
\end{adjustbox}
\caption{\small{{Ablation studies on the discriminator (3D Euclidean distance (mm)).}}}
\vspace{-0.2in}
\label{tab:discriminator}
\end{table}

\textbf{On loss functions.}
\begin{table}[t]
\centering
\begin{adjustbox}{width=\linewidth}
\begin{tabular}{c|ccc|ccc|ccc}
\toprule[1pt]
\multirow{2}{*}{Model} & \multirow{2}{*}{$\mathcal{L}_{\text{pose}}+\mathcal{L}_{\text{proj}}$} & \multirow{2}{*}{$\mathcal{L}_{\text{len}}$} & \multirow{2}{*}{$\mathcal{L}_{\text{dir}}$} & \multicolumn{3}{c}{STB}     & \multicolumn{3}{c}{RHD}     \\ \cline{5-10} 
                       &                     &                     &                     & Stage \RNum{1} & Stage \RNum{2} & Stage \RNum{3} & Stage \RNum{1} & Stage \RNum{2} & Stage \RNum{3} \\ \midrule[0.5pt]
1                      & \checkmark                   &                     &                     & 32.75 & 9.11  & 5.35  & 99.24 & 25.96 & 15.07 \\
2                      & \checkmark                   & \checkmark                   &                     & 30.32 & 8.00  & 5.02  & 95.19 & 22.96 & 14.76 \\
3                      & \checkmark                   &                     & \checkmark                   & 27.65 & 6.91  & 5.00  & 89.76 & 21.63 & 14.01 \\
4                      & \checkmark                   & \checkmark                   & \checkmark                   & 24.15 & 5.12  & \textbf{3.97}  & 83.37 & 15.84 & \textbf{12.40} \\ \bottomrule[1pt]
\end{tabular}
\end{adjustbox}
\caption{\small{{Ablation studies on the proposed bone-constrained loss functions at three stages.}}}
\label{tab:loss}
\vspace{-0.4in}
\end{table}
We also evaluate the proposed bone-constrained loss functions $\mathcal{L}_{\text{len}}$ and $\mathcal{L}_{\text{dir}}$ separately. 
We train the network with different combinations of loss functions on the STB and RHD datasets in three stages respectively. 
As reported in Tab.~\ref{tab:loss}, the network trained with our proposed bone-constrained loss functions performs better in all the three stages on both datasets. 
We also notice that $\mathcal{L}_{dir}$ plays a more significant role compared to $\mathcal{L}_{\text{len}}$.
This gives credits to the constraint on the orientation of bones that explicitly takes structural properties of hands into consideration.


\vspace{-0.1in}
\section{Conclusion}
\label{sec:conclude}
In this paper, we propose regularized graph representation learning under a conditional adversarial learning framework for 3D hand pose estimation from a monocular image. 
Based on the MAP estimation formulation, we take an initial estimation of hand pose as prior pose, and further learn the structural deformation in the prior pose via residual graph convolution. 
Also, we propose two bone-constrained loss functions to enforce constraints on the bone structures explicitly. 
Extensive experiments demonstrate the superiority of the proposed method.  

%
%
%
\bibliographystyle{splncs04}
\bibliography{ref}
\end{document}


\maketitle

In this supplementary material, we present more experimental results, including: 
\begin{itemize}
    \item the results of ablation studies in the evaluation metric of {\it 3D PCK};
    \item the comparison results with state-of-the-art methods in the evaluation metric of {\it 3D PCK}; 
    \item more qualitative results. 
\end{itemize}

\section{Ablation Study Results in 3D PCK}
Regarding the ablation studies in Section 4.3 in the paper, while we have presented the results in {\it 3D Euclidean distance (mm)} in the paper, we further show the results in {\it 3D PCK}, including results on different stages, on the deformation learning module, on the conditional discriminator, and on loss functions in Fig.~\ref{fig:stage}, Fig.~\ref{fig:ref}, Fig.~\ref{fig:dis} and Fig.~\ref{fig:loss}, respectively.

\section{Comparison to the State-of-the-Art in 3D PCK}
In addition to the comparison results in 3D Euclidean distance (mm) in Section 4.4 of the paper, we provide the results in 3D PCK on the RHD dataset \cite{Zimmermann_2017_ICCV}, the MPII+NZSL dataset \cite{Simon_2017_CVPR}, the EGODEXTER dataset \cite{OccludedHands_ICCV2017} and the Dexter+Object dataset \cite{OccludedHands_ICCV2017}, in comparison with \cite{Baseline}, \cite{Wild}, \cite{Cai_2018_ECCV}, \cite{Spurr}, \cite{Zimmermann_2017_ICCV}, \cite{Iqbal} and \cite{GANeratedHands_CVPR2018} in Fig.~\ref{fig:RHD}, Fig.~\ref{fig:MPII}, Fig.~\ref{fig:EGODEXTER} and Fig.~\ref{fig:Dexter+Object} respectively.

\section{More Qualitative Results}
In addition to the qualitative results in Fig.~\ref{fig:vis} of the paper, we further present more qualitative results on the RHD dataset and the MPII+NZSL dataset in Fig.~\ref{fig:visrhd} and Fig.~\ref{fig:vismpii} respectively in the supplementary material.

{
\bibliography{ref_supp}
}

\begin{figure*}
    \centering
    \begin{minipage}[t]{\linewidth}
    \includegraphics[width=\linewidth]{supp_figs/Ablation Stage.png}
    \caption{Ablation study on different stages in 3D PCK. }
    \label{fig:stage}
    \end{minipage}
    \\[1pt]
    \begin{minipage}[t]{\linewidth}
    \includegraphics[width=\linewidth]{supp_figs/Ablation Refinement.png}
    \caption{Ablation study on the deformation learning module in 3D PCK.}
    \label{fig:ref}
    \end{minipage}
    \\[1pt]
    \begin{minipage}[t]{\linewidth}
    \includegraphics[width=\linewidth]{supp_figs/Ablation Discriminator.png}
    \caption{Ablation study on the conditional discriminator in 3D PCK.}
    \label{fig:dis}
    \end{minipage}
    \\[1pt]
    \begin{minipage}[t]{\linewidth}
    \centering
    \includegraphics[width=0.7\linewidth]{supp_figs/Bone Loss Ablation.png}
    \caption{Ablation study on loss functions in 3D PCK. }
    \label{fig:loss}
    \end{minipage}
\end{figure*}

\begin{figure*}[t]
    \centering
    \begin{minipage}[t]{0.42\linewidth}
    \includegraphics[width=\linewidth]{supp_figs/PCK RHD.png}
    \caption{3D PCK on RHD dataset}
    \label{fig:RHD}
    \end{minipage}
    \quad 
    \begin{minipage}[t]{0.5\linewidth}
    \includegraphics[width=\linewidth]{supp_figs/result MPII.png}
    \caption{3D PCK on MPII+NZSL dataset}
    \label{fig:MPII}
    \end{minipage}
    \\[1pt]
    \begin{minipage}[t]{0.44\linewidth}
    \includegraphics[width=\linewidth]{supp_figs/PCK EGODEXTER.png}
    \caption{3D PCK on EGODEXTER dataset}
    \label{fig:EGODEXTER}
    \end{minipage}
    \quad
    \begin{minipage}[t]{0.44\linewidth}
    \includegraphics[width=\linewidth]{supp_figs/PCK Dexter+Object.png}
    \caption{3D PCK on Dexter+Object dataset}
    \label{fig:Dexter+Object}
    \end{minipage}
\end{figure*}



\begin{figure*}[h]
\centering
 
\subfigure[Qualitative Results on RHD]{
    \begin{minipage}[t]{0.47\linewidth}
        \centering
        \includegraphics[width=\linewidth]{supp_figs/RHDvis.png}
        \vspace{0.02cm}
        \label{fig:visrhd}
    \end{minipage}%
}%
\subfigure[Qualitative Results on MPII+NZSL]{
    \begin{minipage}[t]{0.47\linewidth}
        \centering
        \includegraphics[width=\linewidth]{supp_figs/MPIIvis.png}
        \vspace{0.02cm}
        \label{fig:vismpii}
    \end{minipage}%
}%
\caption{Qualitative Results on the RHD dataset and the MPII+NZSL dataset from Our Method.}
\label{fig:vis}
\end{figure*}